# The effect of fatigue on the performance of online writer recognition

Enric Sesa-Nogueras, Marcos Faundez-Zanuy[*] and Manuel-Vicente Garnacho-Castaño
Escola Superior Politècnica & Escola Superior de Ciències de la Salut
TecnoCampus Mataró-Maresme

**Abstract**

Background: The performance of biometric modalities based on things *done by the subject*, like signature and text-based recognition, may be affected by the subject's state. Fatigue is one of the conditions that can significantly affect the outcome of handwriting tasks. Recent research has already shown that physical fatigue produces measurable differences in some features extracted from common writing and drawing tasks. It is important to establish to which extent physical fatigue contributes to the intra-person variability observed in these biometric modalities and also to know whether the performance of recognition methods is affected by fatigue. Goal: In this paper we assess the impact of fatigue on intra-user variability and on the performance of signature-based and text-based writer recognition approaches encompassing both identification and verification.
Methods: Several signature and text recognition methods are considered and applied to samples gathered after different levels of induced fatigue, measured by metabolic and mechanical assessment and, also by subjective perception. The recognition methods are Dynamic Time Warping and Multi Section Vector Quantization, for signatures, and Allographic Text-Dependent Recognition for text in capital letters. For each fatigue level, the identification and verification performance of these methods is measured. Results: Signature shows no statistically significant intra-user impact, but text does. On the other hand, performance of signature-based recognition approaches is negatively impacted by fatigue whereas the impact is not noticeable in text-based recognition, provided long enough sequences are considered.

## 1. Introduction

Signature-based and, to a lesser extent, text-based writer recognition are popular biometric modalities based on behavioural traits: something an individual can do. This family of biometric modalities possess some relevant advantages, such as the impossibility of being lost, stolen, or shared and its non-invasiveness, but also suffers from some well-known drawbacks, like the possibility of being faked, the variation over time (e.g., aging) and the variation due to the individual's state. The latter drawback can in some circumstances be considered a positive property since information regarding a person's state can be deduced from the comparison of different samples of their handwriting. This handwriting-based assessment has been applied to the early diagnose of neurodegenerative conditions [1] [2] [3], to the measurement of the effect of oxygen therapy in patients suffering from chronic obstructive pulmonary disease [4] and to assess the effects of alcohol [5] [6], marijuana [7] or caffeine [8].

---

[*] Corresponding author: faundez@tecnocampus.cat

Fatigue, the physical sensations experienced by an individual when incurring in physical activity, is one of the states that can impact the handwriting of any person. Physical fatigue has effects on the perceptual and motor machinery of the human body which translate into variations in the output produced. Calligraphic experts are perfectly aware that fatigue may change some properties of any person's handwriting and that in some cases these changes may be detected by visual inspection [9]. This issue has received some attention from the field of forensics. In [10] it is reported that under fatigue there is an increase in vertical height and in letter width, a slight deterioration in writing quality (carelessness, scrawling), omission of punctuation and diacritics and the enlargement of minute movements. [9] also reports on a study conducted by Remillard [11] linking pulse rate and handwriting quality deterioration, assessed by calligraphic experts. One conclusion in [11] pointed out that the degradation induced by physical fatigue was similar, yet less pronounced, to the degradation following alcohol consumption. In [12] the authors also concluded that fatigue induced by the repetition of a writing task has observable effects even on control subjects not suffering from movement disorders.

If fatigue has an impact on the features of the handwriting produced by an individual, then it follows that it may also have an impact on the performance of recognition methods based on them. Given two samples, the matching engine of the recognition system measures the distance (dissimilarity) between them and then transforms it into a score that quantifies their degree of similarity. In verification, the samples will be deemed having been produced by the same person if the score is above a given threshold. In identification, the authorship of a sample is granted to the writer whose model yields a higher score. Two sample achieving the highest theoretical score are very unlikely to occur due to intra-subject and inter-subject variability. The former is responsible for false non-matches (false rejections in verification) while the latter is responsible for false matches (false acceptances in verification). Figure 1 graphically depicts these concepts. In statistics, they are usually referred as type-I and type-II errors, respectively. Intra subject variation is of relevance in biometric recognition since it establishes the *normal* range of variation that individuals exhibit. Any situation increasing the intra-subject variability will also increase the probability of a false non-match.

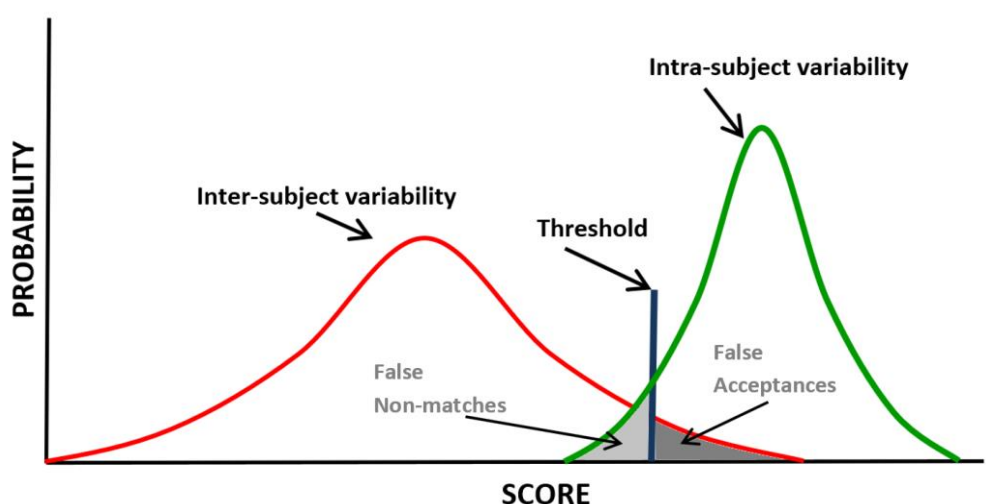

**Fig. 1: Intra-Subject and Inter-subject-variability.**

Several measures have been proposed to summarize the performance of a recognition system. Throughout this paper the equal error rate (EER) will be used for verification while the identification rate (IDR) will be used for identification. EER is the value that satisfies that the false acceptance rate (FAR) equals the false rejection rate (FRR). The lowest this value the better the performance (overall accuracy) of the verification method. IDR is the ratio of well identified subjects expressed as a percentage. The highest this value the better the performance of the identification system. Notice that while EER is independent of the number of individuals considered, IDR is not, since it tends to decrease as the number of individuals grows.

## 2. Recognition methods

### 2.1. *Signature-based recognition*

For signature recognition two different methods have been used: Dynamic Time Warping (DTW) and Multi-Section Vector Quantization (MSVQ).

DTW is a well-known template matching technique well suited to cope with random variations due to the writer's behaviour (pauses, hesitations) [13]. DTW applies a dynamic programming strategy to produce a measure of the distance between two samples even if they differ in length. It has been widely used in signature-based writer recognition.

```
INPUT:
        S₁ a sequence of l₁ elements indexed in [1… l₁]
        S₂ a sequence of l₂ elements indexed in [1… l₂]
OUTUPT
        A measure of the dissimilarity between s₁ and s₂

distMat a two-dimensional matrix indexed in [0… l₁][0… l₂]

for all i, j in [0…l₁] and [0… l₂] respectively
        distMat[i,j] = +∞

distMat[0,0] = 0.0

for all i in [0, l₁]
        for all j in [0, l₂]
                cost = distance between S1[i] and S2[j]
                distMat[i, j] = cost + MINIMUM(
                                        distMat[i-1, j],
                                        distMat[i, j-1],
                                        distMat[i-1, j-1])

result is distMat[l₁, l₂]
```

**Fig. 2:** DTW algorithm.

We successfully applied DTW to signature identification and verification in [14]

MSVQ was proposed in the eighties for speech recognition and was at that time named multi-section codebook [15] [16].

Although this approach was discarded in speech recognition due to the higher accuracies of Hidden Markov Models (HMM) and some other more complex techniques, it has proven adequate in signature-based recognition where training sets contain a reduced number of shorter samples. Contrary to HMM and other statistical approaches, MSVQ shows acceptable accuracies with small training sets. In [17] and [18] we reported on successful applications of MSVQ to signature.

MSVQ starts by splitting the training samples into several sections (e.g., in a three-section approach, the approach used in the experimental section, each signature is split into three equal-length parts: the initial, middle and final sections). Then a quantizer, in the form of a b-bit codebook, is generated for each user and section applying the LBG algorithm [19]. These codebooks constitute the user's model. In order to take a decision on a sample, it is encoded using the user's model (their codebooks) and the quantization distortion is used as a measure of dissimilarity: the higher the distortion the more dissimilar the given sample to the samples that originated the model. Each section contributes a measure and the final measure is obtained by averaging. Identification attributes the authorship of the sample to the user whose model yields less distortion, whereas verification deems the sample as belonging to a user if distortion is below a certain threshold.

### 2.2. *Text-based recognition*

Text-based writer-recognition is a biometric modality that has received much less attention than signature since it is often believed that text and signature are just handwriting products being two sides of the same coin and that short sequences of text possess less discriminative power (less potential for correct writer identification) than signature [20]. However, some authors regard them as different biometric modalities [21]. Being a much less researched modality, it cannot be said that widely used methods exist, much less in the online domain. Although signature approaches can be adapted to be applied to sequences of text, results tend to be less accurate -worst performance- since text shows less inter-subject variability due to legibility constraints: an individual can choose their signature and it may lack legibility while the same does not applies to text.

For text-based recognition the allographic text-dependent recognition system (ATDR) presented in [22] has been applied. This approach yields performances close to those obtained with signature and, with minor variations, it has also been successfully applied to gender recognition [23]. It fully exploits online data, and benefits from the information provided by in-air trajectories, and from the combination of in-air and on-surface signals.
ATDR is a stroke-based schema in which text is regarded as a pair of sequences: one of in-air strokes (invisible, performed while the writing device was not touching the writing surface) and one of on-surface strokes (visible). The system relies in a pair of catalogues of strokes obtained by means of a previously trained Kohonen network (KN) [24]. Each catalogue is set of stroke-prototypes, equivalent to a vector quantizer, that can classify each stroke into a class. For the purpose of the experimentation reported in this paper, the catalogues, one for each word and type of stroke, were built from samples coming from the BiosecurID database [25]. In ATDR each realization of a word is pre-processed (see [22] for further details) and then converted (re-encoded) into a sequence of integers, where each integer represents a stroke-prototype. Recognition is performed by the comparison of re-encoded sequences using DTW and taking advantage of the neighbouring properties of the KN. Two dissimilarity measures are obtained, one for the in-air sequence and another one for the on-surface sequence. In a later stage, these two measures can be combined into a single one. Figure 3 gives a graphical depiction of the process.

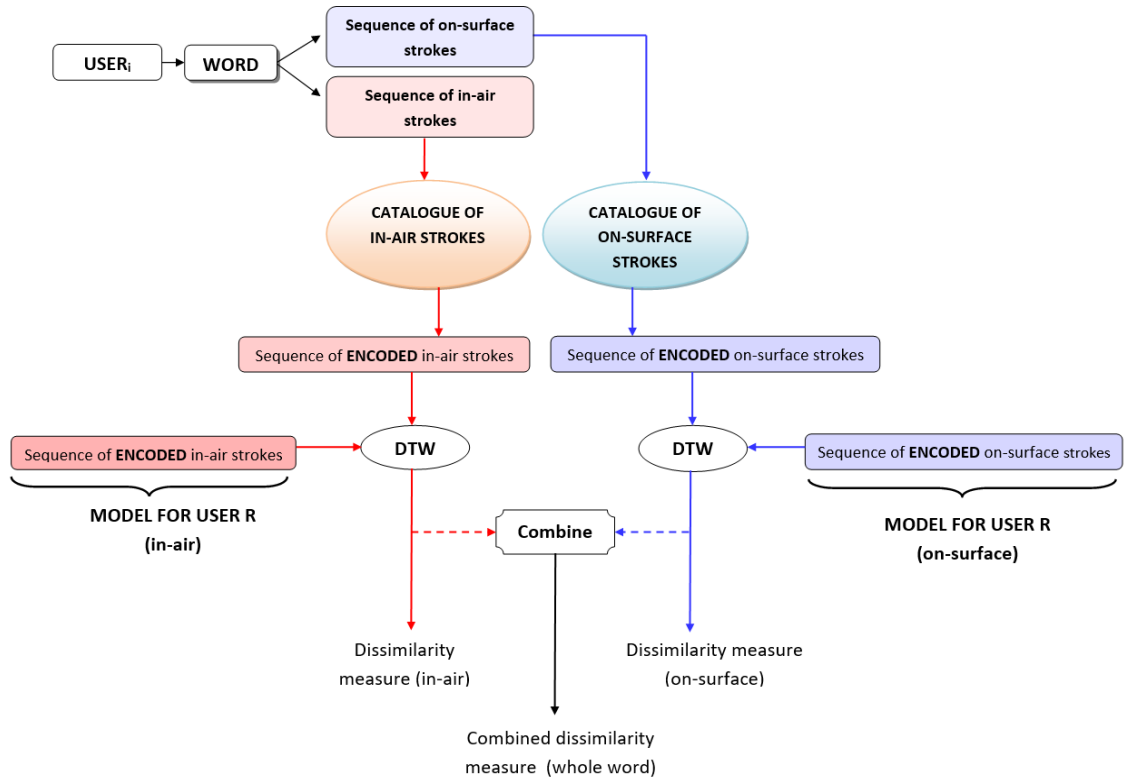

**Fig. 3:** schematic overview of the allographic text-dependent recognition system (ATDR) used for text-based writer recognition

## 3. Experimental results

### 3.1. Database and procedure

All the experimentation reported in this paper has been carried out using data from the *Writing & Fatigue* database presented in [26]. This database was built with the goal of being a tool to study the influence of the fatigue induced by physical exercise in different handwriting tasks, including signature and text. The *Writing & Fatigue* database contains handwritten samples from 20 healthy young males (Refer to [26] for further details regarding the donors' selection criteria). Acquisition was carried out using a WACOM INTUOS tablet. The authors acknowledge that the size of the database is moderately small, but it is, to the best of our knowledge, the only existing database that provides handwriting samples acquired, in a controlled way, under different levels of fatigue. This limitation in the number of individuals stems from the complexity of the building of the database, not only because of the time required to gather each individual's samples, including blood collection, but also because the participants had to accomplish very restrictive criteria regarding their physical condition since they had to perform a strenuous physical activity.

Table 1 summarizes the most relevant statistics regarding the participating donors.

| GENDER | AGE | WEIGHT (Kg) | HEIGHT (cm) | BODY MASS INDEX (Kg·m-2) |
|---|---|---|---|---|
| Male | $\in [18, 24]$ | $\in [64.4, 79.4]$ | $\in [168.4, 182.8]$ | $\in [19.8, -26.6]$ |

**Table 1 Relevant statistics regarding the donors in the *Writing & Fatigue* database**

Handwriting samples were taken:
- After a light warm up that did not induce any noticeable fatigue (BASE samples)
- After a mild exercise to induce a light fatigue (MEIF samples)
- Before starting a strenuous exercise (these samples not considered in this paper)
- After a strenuous exercise to induce a high fatigue level (SEIF samples)
- After a short recovery time (3 minutes) from the previous strenuous exercise (post-SEIF samples)

**Fig. 4: Acquisition sheet with slots for different handwriting tasks**

These samples comprised two executions of the donor's signature (slots 4 and 8 in the acquisition sheet shown in figure 4) and one execution of each of the words in table 2 (slot 6 in the acquisition sheet). These four particular words were used because they also appear in the BIOSECURDId database [25] and because we already possess catalogues of their allographic components [22].

| WORD | TEXT | LENGTH |
|---|---|---|
| W1 | BIODEGRADABLE | 12 |
| W2 | DELEZNABLE | 10 |
| W3 | DESAPROVECHAMIENTO | 18 |
| W4 | DESBRIZNAR | 10 |

**Table 2 Words in the *Writing & Fatigue* database**

Along with the handwriting samples, the fatigue induced by the exercises was also measured. Three types of measures were taken:
- Metabolic: lactate concentration in venous blood.
- Mechanical: vertical flight height [27], [28]
- Subjective: rating of perceived exertion (RPE) using Borg's scale [29]

These three measures give different perspectives of the same phenomenon and allow for a more complete analysis of the interactions among fatigue and recognition performance.

| SCALE | PERCEIVED EXERTION |
|---|---|
| 10 | MAXIMUM EFFORT ACTIVITY. Cannot keep going. Breathless. Cannot talk. |
| 9 | VERY HARD ACTIVITY. Very difficult to keep going. Almost breathless. Can hardly talk. |
| 7-8 | VIGOROUS ACTIVITY. Near uncomfortable. Short of breath. Can speak but only short sentences |
| 4-6 | MODERATE ACTIVITY. Still somewhat comfortable. Breathing heavily. Can talk for a while. |
| 2-3 | LIGHT ACTIVITY. Comfortable. Could keep going for a long time. Can talk normally. |
| 1 | VERY LIGHT ACTIVITY. Almost no exertion at all, just above sleeping or watching TV. |

**Table 3** Borg's scale of perceived exertion expressed in the [1, 10] interval.

For each participant, his BASE samples were used as models (enrolment). Actually:
- Two realizations of the signature.
- One realization of each word.

Then the distances between the models and the rest of samples were determined (testing):

- DTW and three-section MSVQ were applied to MEIF, SEIF and post-SEIF signature samples.
- ATDR was applied to MEIF, SEIF and post-SEIF text samples.

In each case, the recognition performance was calculated. As recognition comprises both identification and verification, two different rates were obtained:
- The rate of well-identified writers (IDR)
- The equal-error-rate (EER): the rate at which false rejections and false acceptances are equal.

Distances themselves were also analysed in order to determine whether they show statistically significative differences depending on their origin (MEIF, SEIF and post-SEIF phase).

*3.2. Signature*

Signatures were compared using DTW and three-section MSVQ. For the former, the distances among the models and the samples were collapsed into a single measure by either taking the minimum one or taking the average. Both measures are shown in the tables and plots that follow. For the latter, models of different sizes (number of bits, from 2 to 7) were considered. As differences were very small, it was decided that only results for 3 and 6-bit models would be considered and plotted, this allowing for less cluttered tables and plots.

Table 4 summarizes the results obtained when intra-user distances obtained in the different phases were compared to determine how different they are depending on the phases they were taken.
First of all, distances were tested for normality putting them to a Lilliefors normality test. For MSVQ, distances from all phases were found incompatible with normality (p-value < 0.001 in all cases) while for DTW only the MEIF distances could be deemed compatible with normality (p-value > 0.3 for both min and mean). With most distances being incompatible with normality, using a Student's t-test was ruled out.
Then distances were put to a Wilcoxon signed-rank test with null hypothesis $H_0$: *distances come from populations with no different means*.

| | Recognition method | | | | | | | |
|---|---|---|---|---|---|---|---|---|
| | DTW | | | | MSVQ | | | |
| | min | | mean | | 3-bit | | 6-bit | |
| Phase | SEIF | Post-SEIF | SEIF | Post-SEIF | SEIF | Post-SEIF | SEIF | Post-SEIF |
| MEIF | 0.0872 | 0.314 | 0.247 | 0.268 | 0.291 | 0.379 | 0.118 | 0.159 |
| SEIF | | 0.513 | | 0.9 | | 0.92 | | 0.96 |

**Table 4 Statistical significance (p-value) of the difference in intra-user distances**

In no case the p-value obtained is lower than a significance level of $\alpha=0.05$ hence there is no strong statistical evidence that fatigue has a significant impact on the measured intra-user distance. Nevertheless, it is worth noticing that p-value is much lower when comparing phases with clearly different fatigue levels (e.g. the p-value for MEIF vs SEIF is much lower than the p-value for SEIF vs Post-SEIF). It is also worth noticing that with a less restrictive significance level ($\alpha=0.1$) DTW with min would have been deemed as showing a significant difference in the MEIF vs SEIF case (p-value=0.0872).

Although the comparisons of intra-user distances do not let us claim the existence of a strong statistical evidence pointing towards a significant impact of fatigue, the performance of all the recognition methods considered does show an affect that can be attributed to fatigue. Table 5 summarizes the results regarding performances (Plotted in figures 5 and 6)

|  | | Recognition method | | | | | | | |
|---|---|---|---|---|---|---|---|---|---|
| Phase | Fatigue indicators | | | DTW | | | | MSVQ | | | |
| | | | | IDR (%) | | EER (%) | | IDR (%) | | EER (%) | |
| | Lactate | MFFH | RPE | min | mean | min | mean | 3-bit | 6-bit | 3-bit | 6-bit |
| BASE | 1.11 | 36.12 | 1.13 | | | | | | | | |
| MEIF | 1.01 | 35.21 | 1.82 | 100 | 100 | 4.76 | 4.94 | 97.62 | 95.23 | 9.94 | 7.14 |
| SEIF | 14.19 | 33.25 | 8.32 | 92.86 | 92.86 | 7.14 | 7.44 | 90.48 | 92.85 | 13.99 | 9.05 |
| post-SEIF | 10.49 | 34.09 | 5.26 | 95.23 | 97.61 | 7.32 | 7.14 | 90.48 | 92.85 | 14.34 | 11.07 |

**Table 5 Summary of performance results for signatures**

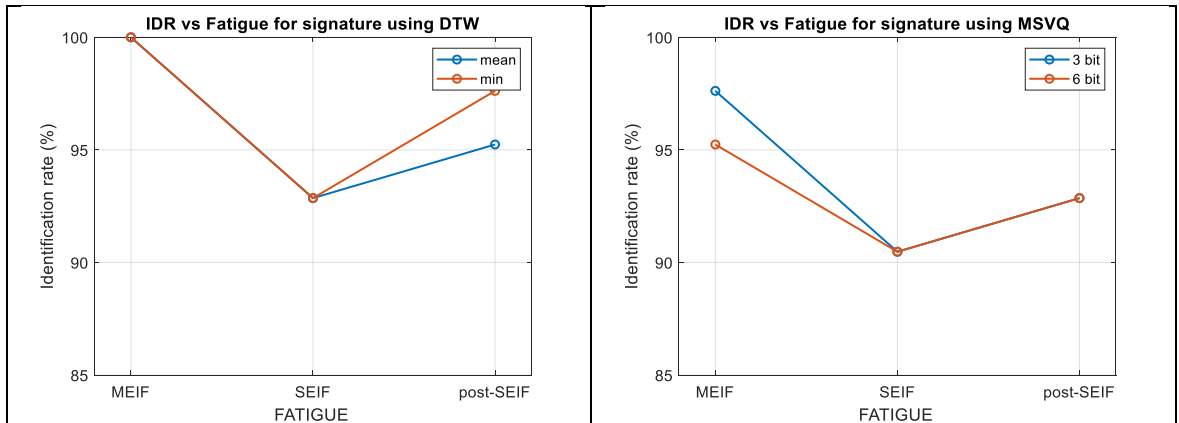

**Fig. 5: Identification performances yielded by MEIF, SEIF and post-SEIF samples.**

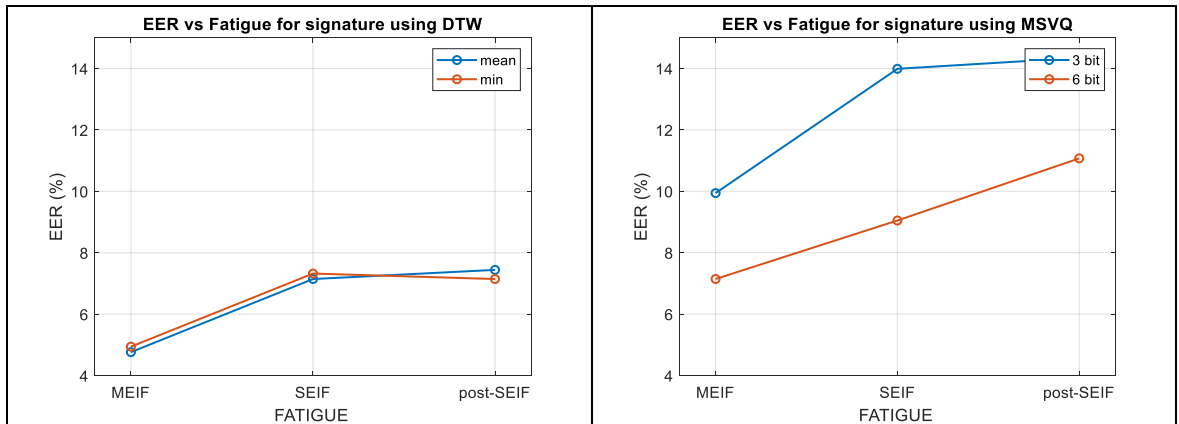

**Fig. 6: verification performances yielded by MEIF, SEIF and post-SEIF samples.**

As for the identification rate, both methods quite clearly suggest that fatigue accounts for worse rates and that the effects start to reverse after a short rest (V-shaped plot).

When it comes to verification, extreme fatigue seems to substantially degrade performance, but now a short period of rest does not seem enough to reverse the decrease in accuracy. What is more, the results might indicate that the effects of fatigue continue to grow at least for some minutes after the strenuous exercise has finished. Hence recovery does not have the same effect on identification and verification, with more rest needed to restore verification accuracy than to restore identification accuracy.

For both IDR and EER the results, (see figures 7, 8 ,9, 10, 11, and 12) suggest that it is possible to draw the same conclusions when the actual measures of fatigue (lactate, flight height and RPE) are considered.

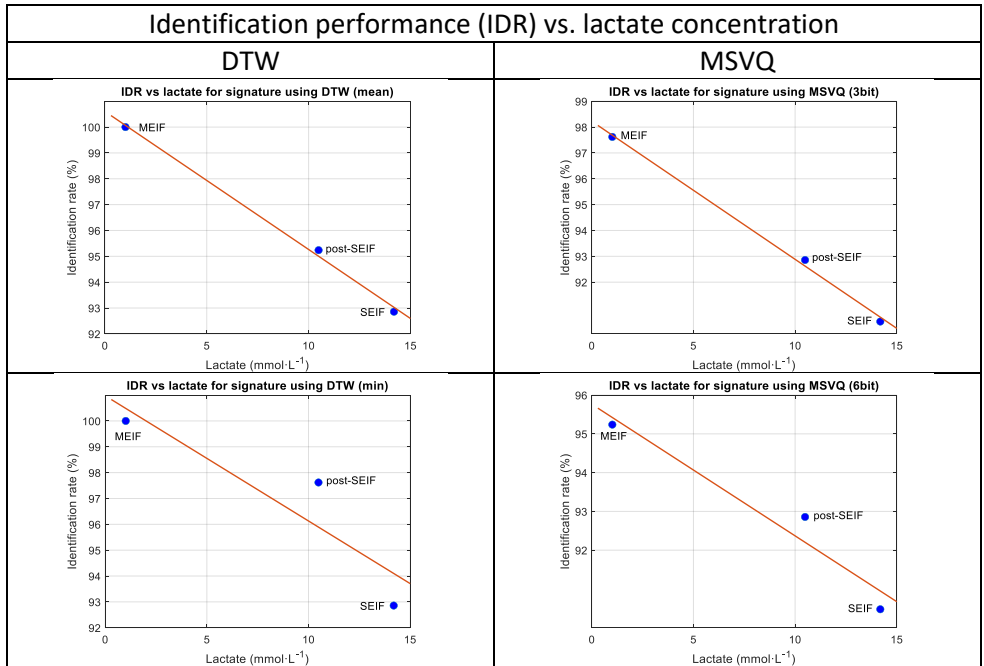

**Fig. 7: identification performances vs. metabolic fatigue measured through lactate concentration for DTW (left) and MSVQ (right).**

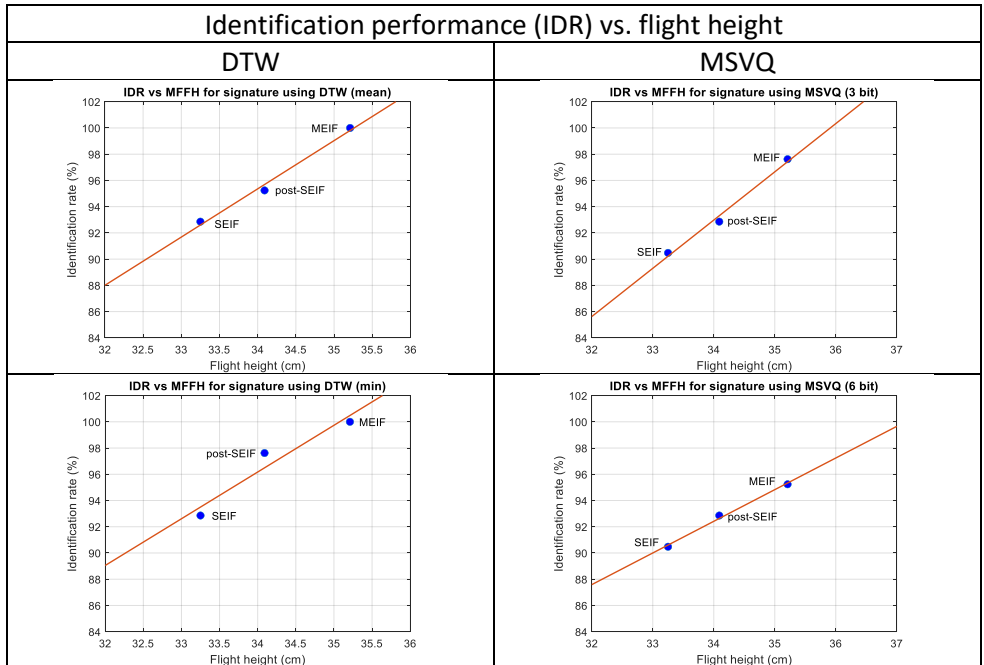

**Fig. 8: identification performances vs. mechanical fatigue measured through flight height for DTW (left) and MSVQ (right).**

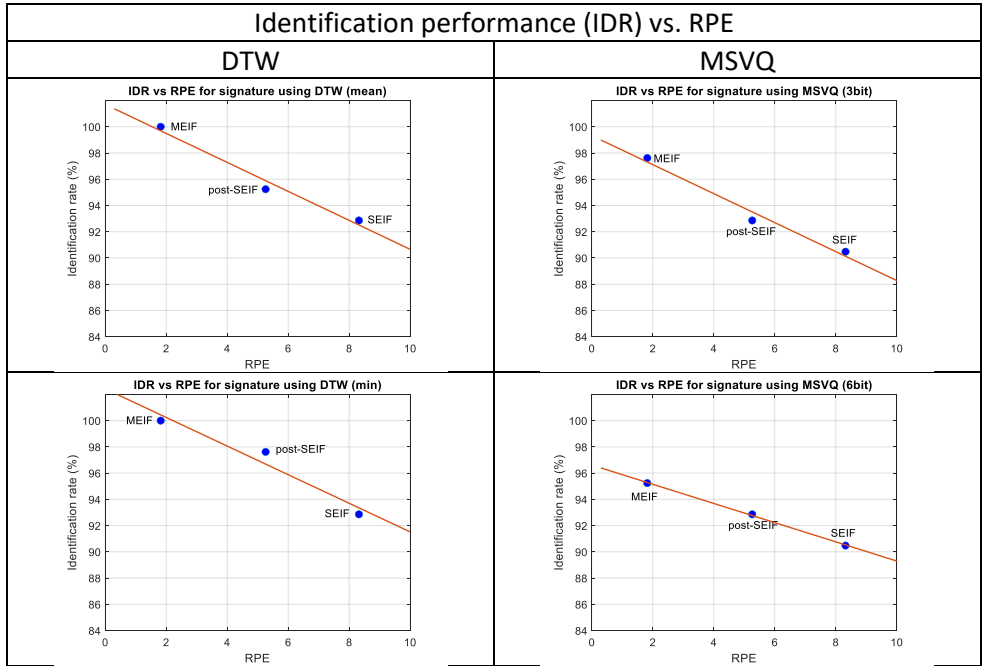

**Fig. 9: identification performances vs. subjective fatigue measured by RPE for DTW (left) and MSVQ (right).**

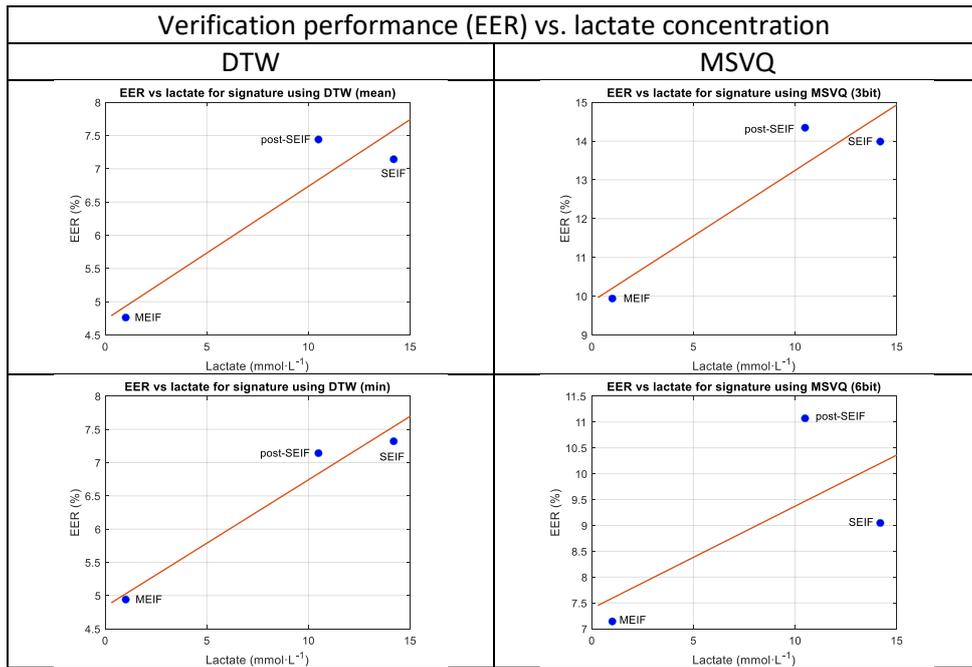

**Fig. 10: verification cation performances vs. metabolic fatigue measured through lactate concentration for DTW (left) and MSVQ (right).**

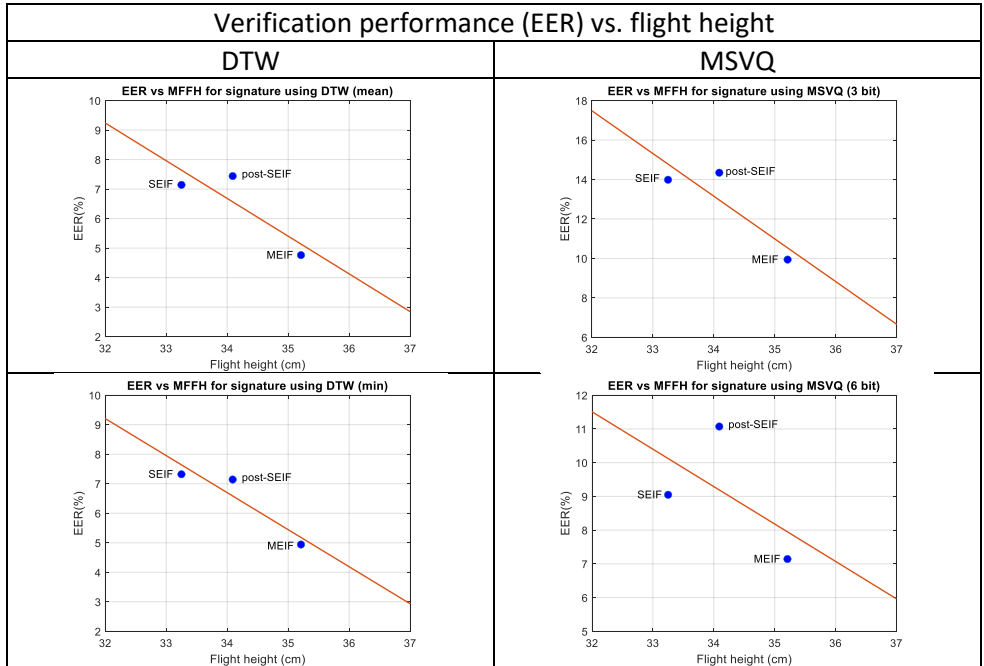

**Fig. 11:** verification performances vs. mechanical fatigue measured through flight height for DTW (left) and MSVQ (right).

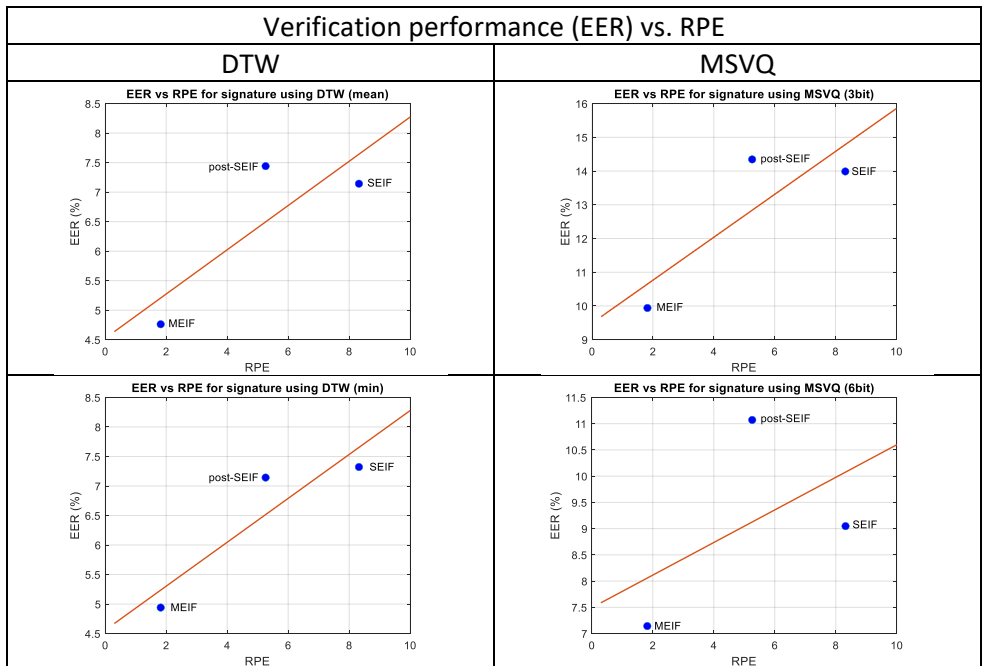

**Fig. 12:** verification performances vs. subjective fatigue measured by RPE for DTW (left) and MSVQ (right).

Regardless of the verification method (DTW or MSVQ) and the type of fatigue measurement (metabolic, mechanical, or subjective) performance decreases after a strenuous exercise and does not improve after a short rest. On the contrary, performance degradation continues even after a short rest. Notice that after the short rest, subjective fatigue dropped from 8.32 to 5.26 in the Borg's scale (from more than "near uncomfortable" to "still somewhat comfortable").

### 3.3. Text

Contrary to signature, text does show, in some cases, a statistically significant difference in the intra-user distances attributable to fatigue (significance level $\alpha=0.05$). Table 6 contains the p-values obtained from the application of a Wilcoxon signed-rank test with null hypothesis $H_0$: *distances come from populations with no different means*. (Similar to signatures, not all intra-user distances can be deemed compatible with normality when put to a Lilliefors test.)

P-values lower than 0.05 have been highlighted. As in its core the ATDR method used in the experiments reported in this section can distinguish between on-surface and in-air trajectories, the results that follow consider these trajectories separately and combined.

|  | In-air | | On-surface | | Both | |
| --- | --- | --- | --- | --- | --- | --- |
|  | SEIF | Post-SEIF | SEIF | Post-SEIF | SEIF | Post-SEIF |
| MEIF | **0.011** | 0.057 | 0.279 | 0.158 | **0.022** | **0.035** |
| SEIF |  | 0.74 |  | 0.84 |  | 0.86 |

**Table 6 Statistical significance (p-value) of the difference in intra-user distances for text (all four words). In green p-values below $\alpha=0.05$.**

When both in-air and on-surface trajectories are considered, there is a significant difference between MEIF and SEIF distances and also between MEIF and Post-SEIF distances. This significant difference can be interpreted as a noticeable effect of fatigue on the intra-user measures: fatigue changes how words are written, and the change has a statistically significant impact on the measure of their dissimilarity.

There is also a statistically significant difference between MEIF and SEIF distances in the case of in-air trajectories, and when it comes to the difference between MEIF and Post-SEIF the p-value is quite low and just slightly above the significance level.

On the other hand, the impact of fatigue on on-surface trajectories is not statistically significant with MEIF vs SEIF p-values similar to those obtained for signatures.

As in the case of signature, fatigue seems to have an impact on the performance of the recognition system. Table 7 summarize the results obtained and figures 13 and 14 graphically depict them.

| Phase | WORD | | | | | | | | ALL FOUR WORDS AS A WHOLE | | | | | |
| --- | --- | --- | --- | --- | --- | --- | --- | --- | --- | --- | --- | --- | --- | --- |
|  | BIODEGRADABLE | | DELEZNABLE | | DESAPROVECHAMIENTO | | DESBRIZNAR | | In-air | | On-surface | | Both | |
|  | IDR | EER | IDR | EER | IDR | EER | IDR | EER | IDR | EER | IDR | EER | IDR | EER |
| BASE |  |  |  |  |  |  |  |  |  |  |  |  |  |  |
| MEIF | 95.24 | 4.76 | 85.71 | 9.05 | 100 | 3.69 | 85.71 | 4.76 | 100 | 0.71 | 100 | 4.64 | 100 | 3.69 |
| SEIF | 85.71 | 9.52 | 90.48 | 5.48 | 100 | 8.57 | 76.19 | 9.4 | 95.24 | 3.69 | 80.95 | 9.52 | 100 | 4.52 |
| post-SEIF | 80.95 | 9.52 | 90.48 | 9.52 | 95.24 | 8.81 | 76.19 | 14.29 | 100 | 9.17 | 90.47 | 9.52 | 100 | 4.76 |

**Table 7 Summary of recognition performance for text**

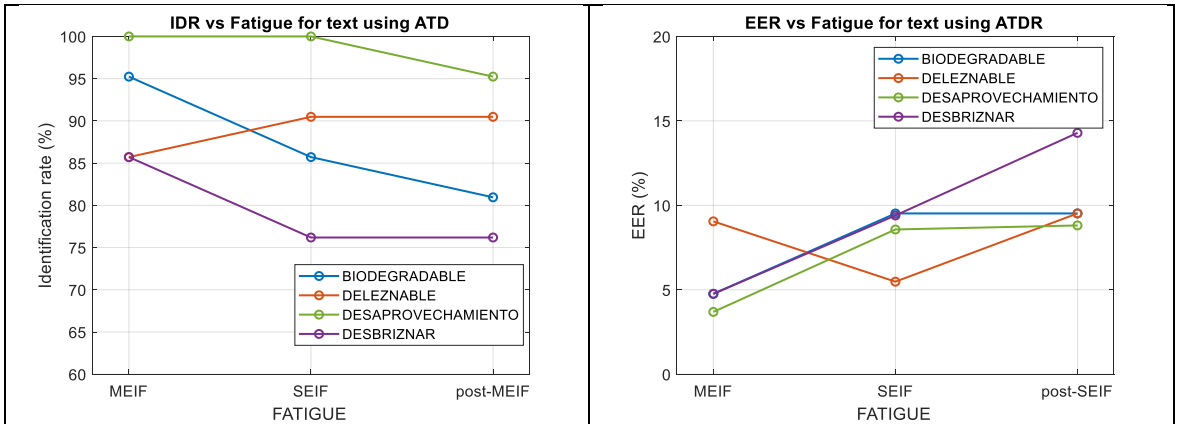

**Fig. 13: recognition performances yielded by MEIF, SEIF and post-SEIF samples when using ATDR and a single word.**

Regarding accuracy, text-based writer recognition shows figures compatible with the results reported for signature-based recognition: it tends to decrease when fatigue increases (3 out of 4 words suffer a reduction). Identification rate diminishes while verification error increases. But this biometric modality is also affected by the particular text used, with one of the words under consideration, DELEZNABLE, yielding an increase in accuracy with higher fatigue levels. This apparent contradiction may be caused by the general lack of accuracy that shorter words manifest [22] [20]

Regarding verification accuracy, it decreases with fatigue (again word DELEZNABLE is the exception) and shows no increase, or continues to decrease, after a short rest.

When all four words are considered as a whole, taking into account both in-air and on-surface trajectories, there's no effect on IDR (see figure 14) while EER shows a very similar pattern to the one obtained for signature.

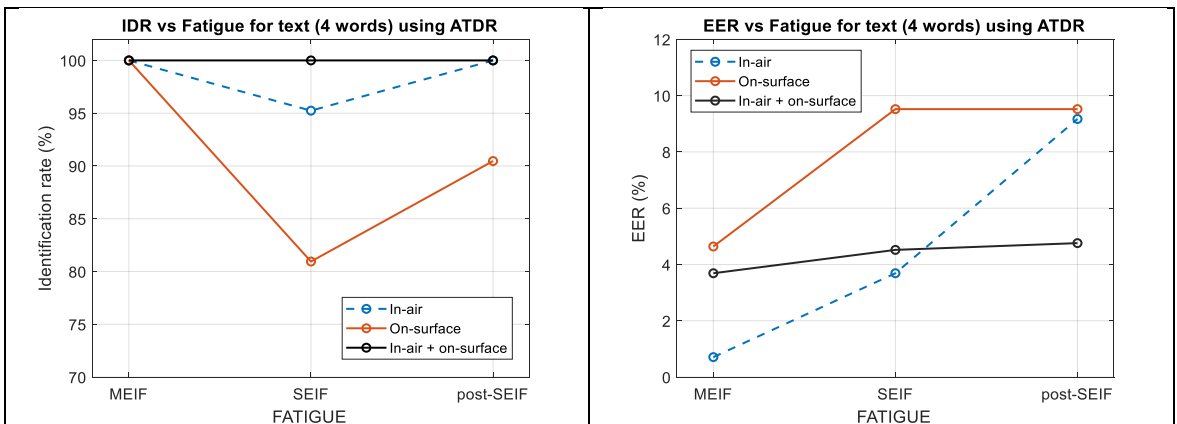

**Fig. 14: recognition performances yielded by MEIF, SEIF and post-SEIF samples when using the four words as a whole (in-air, on-surface and their combination are plotted separately).**

Figure 15 shows how the actual measure of fatigue (lactate, flight height and RPE) relate to verification accuracy (EER). No similar plots are given for identification accuracy (IDR) since its value is 100% in all phases.

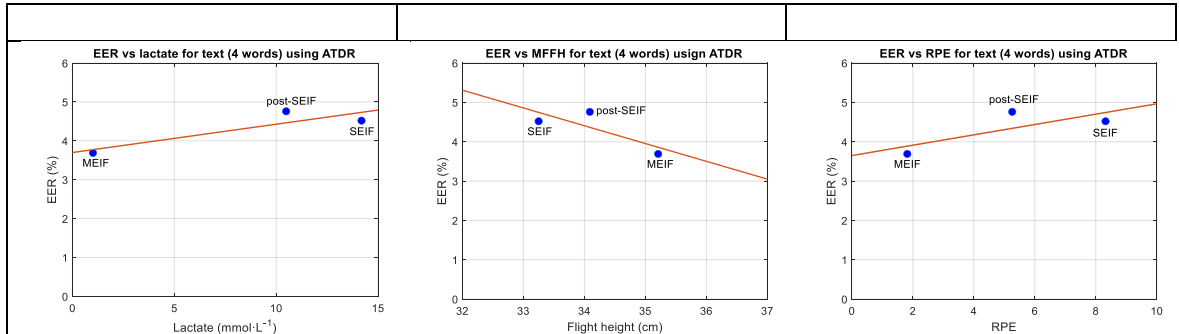

**Fig. 15: verification performance vs fatigue measured by lactate (left), flight height (centre) and RPE (right)**

## 4. Discussion and conclusions

### 4.1. Signature

Intra-user distances taken after a mild exercise inducing a light yet noticeable fatigue (MEIF) cannot be said to be significantly different to the distances taken after an strenuous exercise (SEIF) although both recognition methods analysed, DTW and MSVQ, obtain p-values close to the significance (0.0872 and 0.118).

Signatures are very short writing realizations performed in a quite automatic way and this may explain why the difference in the writer's state is not enough to generate substantial intra-person variabilities. Other explanations may concur, like the accuracy of the approach used to compute the distances. In our experiment, DTW taking the minimum seems to be more sensitive to the effects of fatigue (lower p-values) than the other approaches. This is in agreement with the fact that DTW is more sensitive to mismatches between training and testing conditions, while methods like MSVQ are coarser tending to generalize better at the expense of lower accuracies.

On the other hand, the overall variations in distance, now including the inter-user, have a noticeable effect on the performance of the recognitions systems: when writers are modelled in repose, performance degrades if recognition is done in a high fatigue state: identification rate (IDR) decreases after strenuous exercise and plots suggest that it starts to improve after a short rest, when the fatigue level has descended (see figure 5). When it comes to verification, experimental results point towards a noticeable increase in the overall error although in this case plots do not suggest that a short rest is enough to induce an improvement in the equal error rate (EER).

All three fatigue measures (lactate concentration, flight height and rating of perceived exertion) show similar influences in performance, especially clear in the case of identification.

### 4.2. Text

Text shows results dependent on the length of the sequences considered. With a single word, results are not far from those obtained with signatures: intra-user distances do not show any statistically significant variation, while recognition performance diminishes as fatigue increases. But when a longer sequence is tested, the effect of fatigue in intra-user distances turns out to be statistically significant (with α=0.05). In-air trajectories appear to be more affected than on-surface strokes although the combination of both types of trajectories retains the statistical significance and improves it in the MEIF vs. Post-SEIF case (see table 6). As for why text shows a statistical significance that signature lacks, several explanations may concur: on the one hand the sequences are longer and on the other the writing of words is less automatic than the writing of one's signature since the writer first memorizes the word and then executes it stroke by stroke in a manner that requires the planning of the movements involved.

Regarding the performance of the recognition system used for experimentation, ATDR, short sequences (isolated words) seem to behave like signatures: the identification rate decreases, and the verification error increases. But when it comes to longer sequences - the combination of the four words in our case- the performance remains stable when in-air and on-surface trajectories are combined: IDR is kept to 100% and EER only suffers a slight increment.

**Compliance with Ethical Standards**

The authors declare that they have no conflict of interest.

All procedures performed in studies involving human participants were in accordance with the ethical standards of the institutional and/or national research committee and with the 1964 Helsinki declaration and its later amendments or comparable ethical standards. For this type of study formal consent is not required.

This chapter does not contain any studies with animals performed by any of the authors. Informed consent was obtained from all individual participants included in the study.

**Acknowledgements**

This work was supported by Spanish grant PID2020-113242RB-I00,